# A Hybrid Approach For Hindi-English Machine Translation

Omkar Dhariya*, Shrikant Malviya† and Uma Shanker Tiwary‡
Department of Information Technology
Indian Institute of Information Technology, Allahabad, India 211012
*Email: omkar.dhariya@gmail.com
†Email: shrikant.iet6153@gmail.com
‡Email: ustiwary@gmail.com

*Abstract*—In this paper, an extended combined approach of phrase based statistical machine translation (SMT), example based MT (EBMT) and rule based MT (RBMT) is proposed to develop a novel hybrid data driven MT system capable of outperforming the baseline SMT, EBMT and RBMT systems from which it is derived. In short, the proposed hybrid MT process is guided by the rule based MT after getting a set of partial candidate translations provided by EBMT and SMT subsystems. Previous works have shown that EBMT systems are capable of outperforming the phrase-based SMT systems and RBMT approach has the strength of generating structurally and morphologically more accurate results. This hybrid approach increases the fluency, accuracy and grammatical precision which improve the quality of a machine translation system. A comparison of the proposed hybrid machine translation (HTM) model with renowned translators i.e. Google, BING and Babylonian is also presented which shows that the proposed model works better on sentences with ambiguity as well as comprised of idioms than others.

## I. INTRODUCTION

Cross language communication plays a pivotal role in building a favorable infrastructural environment for multifaceted benefits between two countries. In this internet era, machine translation fulfills the role of an agent to perform this cross language communication. Many countries have put forward enormous efforts for the development of several practical machine translation [1].

Since last few decades, people have tried a significant number of approaches and resources to construct machine translation systems to be utilized in different applications ranging from simple textual translation to multilingual speech systems. Most of the machine translation systems follow the presumption that sentences will be grammatically correct and complete [2] [3]. Then such sentences are translated to target language with preserving the meaning in the source language.

After Mandarin, Spanish and English, Hindi is the most natively spoken language in the world, almost spoken by 260 million people according to Ethnologue, 2014 [4]. Hence, there is vital requirement of many translators capable to translate sentences from Hindi language to other desired target language. We chose English language as target language in this paper. Recently, most of the MT works were focused on English to Indian language translation systems [5], [6]. However, a few systems have been constructed for Hindi to English translation, but not matured enough to resolve all inherent ambiguity and uncertainties of the Hindi sentences.

On the basic level, a machine translator simply converts sentences by substituting word to word from source language to target language. But only the word substitution would not be able to deliver desired results as it doesn't care about semantic and syntactic constraints of the target language. There are many approaches developed to takeover these limitations of automated machine translation such as SMT [7], EBMT [8] and RBMT [9]. These approaches have their own strengths and weaknesses.

There are already existing many freely available Hindi-English machine translation systems like Google Translator, MS-Bing and Babylon. These systems are developed based on different approaches i.e. Rule Based Machine Translation (RBMT), Example Based Machine Translation (EBMT), Statistical Machine Translation (SMT) [6]. But they all not well-accurate in handling the challenges of word sense disambiguation, pronoun resolution and idioms translation.

RBMT systems perform the translation based on the rules discovered by linguists which tell how the words, words sequences or any other structure from source language would be transformed to target language. Other two systems EBMT and SMT extract the rules themselves automatically instead from the parallel corpora developed manually between source and target language. This is why they are referred to as data-driven approaches [5].

Latest approaches of machine translation are the combination of multiple approaches, a "Hybrid Machine Translation". This approach delivers better quality and functionality from traditional approaches [10] [11]. But the problem with HMT is that computationally its more complex than the traditional approaches. A new way of implementing the hybrid approach for machine translation (HMT) has been discussed in this paper that utilize the strength of EBMT, RBMT and SMT. We have also presented the results of experiments performed with our proposed experimental HMT system.

## II. Challenges in Hindi to English Translation

This section reports some inherent challenges in Hindi to English translation systems. Linguistically, morphological manifestation and structural divergences are important characteristics on which the two language can be differentiated. English is based on Subject-Verb-Object (SVO) structure, but Hindi is an Subject-Object-Verb (SOV) type of language. Hindi is morphologically more rich than English. In general, these divergences are the factors which make the translation process difficult and error-prone. Furthermore, Hindi also have some inherent challenges in translating to English (1) Lack of articles in Hindi makes the translation imprecise. (2) Multiple contextual meaning of English prepositions makes it difficult to predict them accurately [12].

The major challenge in the machine translation (MT) between two languages is to identify an inherent translation divergence exist between source and target language. To elaborate more, the divergence can be observed when a sentence in a source language $L_1$ translated to target language $L_2$, in a quite different form [13]. For a robust machine translation (MT) system, it is crucial not only to identify the type of translation divergences but also to resolve them in order to obtain more accurate translation.

In this paper, several Hindi-English translation divergences have been studied in order to identify language specific divergences and further to be incorporated in EBMT and RBMT phases during the translation. In terms of configurational characteristics, English is more rigid and restrictive that follows fixed word order patterns as opposed to Hindi. For instance, one of the translation divergences related to specific word-order pattern is the interpretation of a Hindi question particle "क्या" [14]. "क्या' can be used both as a question particle in *yes/no* type of question sentences and as a type of *interrogative pronoun* in content question sentences as shown in example 1.

**Example 1.** *a. "क्या आप लिख रहे हैं?"* ⇒ ***Are** you writing?*
*b. "आप क्या लिख रहे हैं?"* ⇒ ***What** are you writing?*

Like most South Asian languages, Hindi also shows the replication phenomena of the lexical items to express different grammatical metaphors. In counterpart, English translation doesn't exhibit any such replicative structures. In example 2, the replicative element "चलते चलते" is an adverbial clause which is captured lexically in Hindi. The English counterpart for this example is obtained by a *gerundive propositional phrase*.

**Example 2.** *"वह चलते चलते थक गया।"* ⇒ *He got tired **of** walking*.

The existence of expressive words also responsible to make the translation process error-prone, due to lack of exact parallel counterpart in the target language. Generally, expressive words come from the sound associated with the semantics of the action verb such as टपटपाना (drip), खटखटाना (knock) etc. In example 3, the word धड़ाम is only distantly mapped by 'bump'.

**Example 3.** *"वह धड़ाम से गिरी।"* ⇒ *She fell with **a bump***.

Another source of divergence, in Hindi-English translation is associated with the use of different conjunction and particles in Hindi. In Hindi some of these particles are such as { कि, ना and या etc.} as shown in example 4.

**Example 4.** *"राम स्कूल गया है कि मंदिर।"* ⇒ *Ram went to school **or** temple.*

The above discussion clearly represent the importance of study to exhaust all type of translation divergences in order to build a robust Hindi-English MT. As, it is difficult to process and deal with the all type of translation divergences simultaneously, we have tried to incorporate hybridized example-based and rule-based techniques to sort out some of them.

## III. Approaches for machine translation

Since last several decades, people have developed a number of translation approaches to transform one language content to another ranging from simple word-to-word translation systems to corpus based statistical models as in figure 1. Marcu et al. hypothesizes the translation as :

> "If a sentence to be translated or a very similar one can be found in the TMEM[1], an EBMT system has a good chance of producing a good translation. However, if the sentence to be translated has no close matches in the TMEM, then an EBMT system is less likely to succeed. In contrast, an SMT system may be able to produce perfect translations even when the sentence given as input does not resemble any sentence from the training corpus." [15]

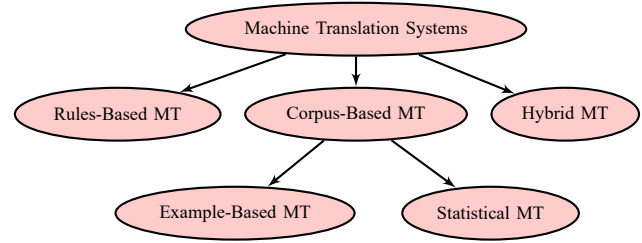

Figure 1. Various approaches of Machine Translations

### A. Rule Based Machine Translation

Jordi et al. have used rule based machine translation for Chinese to Spanish Machine translation [9]. In this work, they have used Apertium platform which is a toolbox for shallow transfer MT. For the generation of translation rules, a bilingual Chinese-Spanish dictionary is constructed consisting of almost 9000 distinctive words. Grammatical transfer-rules were developed manually. They test this system on different domains with average accuracy of 82%.

Pratik et al. have worked on rule based English-to-Hindi machine translation, but their system is domain restricted

---
[1]TMEM = Translation MEMory

[16]. They have used Dependency Parsing as an intermediate representation in the translation. During the translation, the phases of classical analysis, transfer, and generation strategies are replaced with a syntax planning algorithm that directly linearizes the dependency parse of the source sentence as per the syntax of the target language.

## B. Statistical Machine Translation

A statistical approach based English-to-Hindi machine translation system is developed consisting of three processing units Language Model, Translation Model and Decoder [6]. Language model calculates the probability of a sentence in target language. Translation model designed to compute the target sentence probability for the given source sentence. Decoder's job is to select the target sentence which maximizes the probability. The SMT model is trained on the parallel dataset of 5000 sentences pairs. Google translator which is a worldwide renowned and mostly used bilingual translator, is also based on the SMT approach. Google translator learns the SMT parameters from their huge corpus collected from allover the web. The SMT accuracy depends on the corpus quality and the parameter estimation needed be to learn. A group from IBM T.J Watson Reaserch Center work on the Mathematics of SMT and the Parameter Estimation [17].

## C. Example Based Machine Translation

Example based machine translation systems (EBMT) perform the translation of a given input sentence $s$ in three consecutive phases (i) (matching) check for existence of the given input $s$ in the bilingual corpus (ii) (retrieval) extraction of useful segments from the sentence that match in the bilingual corpus and (iii) (transfer) recombining the translated segments [18]. EBMT practically based on the retrieval of source sentences similar to $s$ in the bilingual corpus, hence EBMT is also known as source-similarity based translation. Harold et al. had worked on the Example based machine translation focused on various intuitive problems of the EBMT like the size of Parallel Corpora, Granularity of Examples, Quantity of Examples and Suitability of Examples [8]. Manish et al. proposed to apply EBMT with Fuzzy logic for English to Hindi machine translation. Fuzzy logic implemented in the matching and the alignment of segments during the translation [19].

Table I
MACHINE TRANSLATION ENGINE

| Sr. No. | Translation Engine | Language Support | Multi-Engine Support |
|---|---|---|---|
| 1. | Google[2] | 71 | SMT |
| 2. | MS-Bing[3] | 47 | SMT & RBMT |
| 3. | Babylon[4] | 30 | SMT & Morphological Engine |
| 4. | ImTranslator[5] | 55 | SMT & Other |
| 5. | MyMemory[6] | 151 | SMT |

[2]https://translate.google.co.in/\#hi/en/
[3]https://www.bing.com/translator
[4]http://translation.babylon-software.com/hindi/to-english/

## D. Hybrid Machine Translation

Hybrid machine translation is a method of machine translation that combines characteristics of multiple machine translation approaches within a single machine translation system [10]. Paul et al. worked on a multi-engine hybrid approach to MT, utilizing the statistical models to generate the best possible output from multiple machine translation systems. He has found promising results for Japanese-English machine translation on applying a decision-tree method to select the best possible hypothesis obtained from multiple RBMT, EBMT and SMT decoders [20]. Marcu et al. have also identified the benefits of hybrid MT approaches as coupled multiple MT systems have the precedence over utilizing each MT separately [15]. Evidently, it is been clearly visible that multi-engine MT approaches are capable of surpassing the existing individual MT systems. A comparative study of various existing machine translation with the details of approaches they build on, is shown in table I.

A large number of works have been done on multiple Indian languages i.e. Marathi, Hindi, Sanskrit to English and vice versa at Center for Indian Language Technology (CFILT), IIT Bombay. A Hindi word-net is produced too by CFILT as a semantic relation among the words, which is useful for RBMT systems and helps structural disambiguation to resolve word and attachment ambiguities.

Ondřej Bojar from Charles University in Prague has prepared the Hindi-English parallel Corpus which could be used in the Example Based Machine Translation systems [21].

## IV. PROPOSED APPROACH

### A. Overview

Overall approach has been implemented in a sequence of four primary steps: 1) Segmentation, 2) Translation, 3) POS Tagging and 4) Rearrangement. In figure 2, the flowchart depicts the working relation among multiple steps for translating a Hindi sentence **"विकास विकास ने किया।"** to English **"Vikas did development."**. A inputted sentence has to go through all these steps being transformed from one form to another and at last translated to corresponding English sentence. All these steps are explained briefly in the following subsections.

### B. Algorithm

Algorithm is elaborated in detail in this section through all the steps as in figure 3 with suitable example. The first step in translation is to split the sentence into words or simple sentences (if the input sentence is complex/compound sentence or consists of phrases). If a part of the input is in the example database then we keep that part as it is and the remaining part is segmented into words. When the segmentation is done, the chunks or segments are actually translated and tagged independently. If the segment is example based, it is directly converted to English. Further, the remaining words are translated using the parallel Hindi-English Dictionary

[5]http://imtranslator.net/translation/hindi/to-english/translation/
[6]http://mymemory.translated.net/en/Hindi/English/

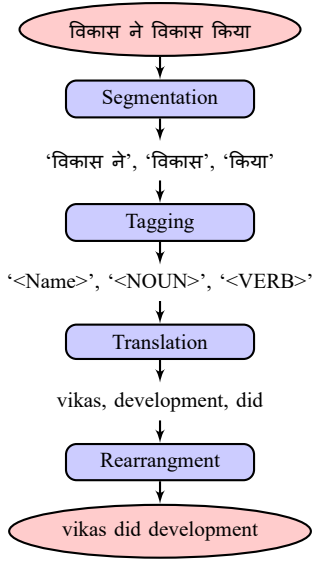

Figure 2. Step-wise result of the purposed approach

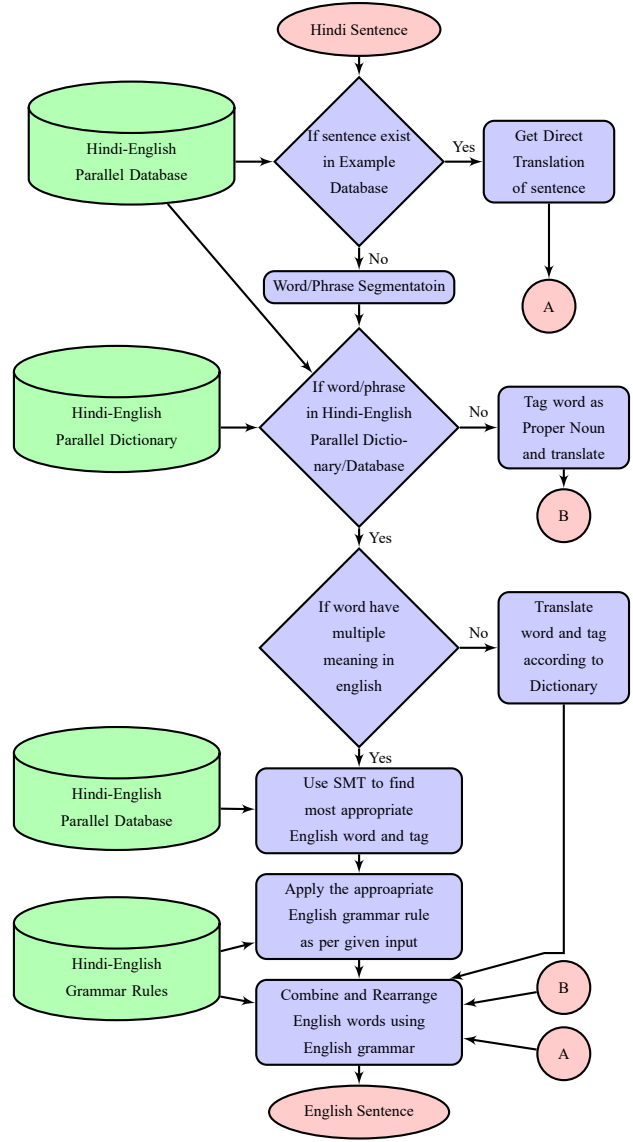

Figure 3. Algorithm of the proposed approach

[22]. Additionally, tagging is also performed at the time of translation with the help of English POS tagger. Meanwhile if a word in Hindi with multiple meaning in English comes into the picture, SMT is used to resolve the ambiguity. SMT calculates the probability of each chunk (word/sentence) based on the probabilistic distribution of it in corresponding source language model to target language model. It's been implemented through Bayes Rule:

$$P_r(S|T) = \frac{P_r(S)P_r(T|S)}{P_r(T)}$$

$P_r(T|S)$: Probability that translator will produce target segment (word/sentence) $T$ when the given source segment is $S$.
$P_r(S)$: Computed by source language model.
$P_r(T)$: Computed by target language model.

Pronouns are identified on the basis of its absence in the Hindi dictionary. Such kind of words are translated through transliteration rules and tagged as pronoun. On finishing the tagging of all the segments, RBMT rearranges them and constructs a valid sentence of it by applying proper grammar.

*1) Segmentation:* Segmentation is performed through first finding all possible sub-parts in the sentence belong to parallel Hindi-English database. Later, the remaining parts of the sentence would be broken into words. At the end of this stage, output would have set of phrases, simple sentences and words. Segmentation helps in making translation process easier in such a way that each segment could be translated separately and combined again to form target sentence as per the EBMT-based translation.

**Example 5.** *Input1:-* *"ओंकार ने मुँह की बात छीनी"*
*Segmentation output1:-* *['ओंकार', 'मुँह की बात छीनी']*
*Input2:-* *"ओंकार और अजय जा रहे थे"*
*Segmentation output2:-* *['ओंकार', 'और', 'अजय', 'जा', 'रहे', 'थे']*

*2) Proper Noun identification:* According to POS, proper noun denotes a particular name used for an individual person, place, or organization. For many languages, a word is known to be as proper noun if it doesn't belong to the dictionary of that language. But in Hindi most of the proper nouns also have the dictionary meaning. This makes the process of pronoun identification more complicated. The problem is solved through incorporating some rules based on the possible contextual morphological information required to denote a word as pronoun as in Example 6.

**Example 6.** *Input1:-* *"मैं ओमकार विकास धारिया हूँ"*
*Tagging1:* *['<PRON>', '<Name>', '<Name>', '<Name>']*
*Input2:-* *"विकास ने विकास किया"*
*Tagging2:-* *['<Name>', '<NOUN>', '<VERB>']*

*3) Tagging:* Tagging is the process to identify the linguistic properties of each individual textual unit. The parallel Hindi-English dictionary contains the tag of each English word. Depending upon the assigned tag, the system finds the proper grammatical structure for the sentence and rearranges the words to construct a grammatically correct sentence. Various type of tags used during the translation are shown in table II.

Table II
TAGS USED IN THE TRANSLATION

| TAG | POS | Details |
|---|---|---|
| PRON | Pronoun | A pronoun is a word that takes the place of a noun. Ex: मैं (I), वह (He/She) |
| ANIMT | ANIMATE Noun | A semantic category of NOUN, referring to a person, animal, or other creature, in contrast to an inanimate noun, which refers to a thing or concept. Ex: लड़के (Boys) |
| VERB | Verb | A verb is a word that in syntax conveys an action, an occurrence, or a state of being (be, exist, stand). Ex: बोलना (To speak) |
| ADJ | Adjective | An adjective is a describing word, the main syntactic role of which is to qualify a noun or noun phrase, giving more information about the object signified. Ex: अच्छा (Good) |
| ADV | Adverb | A word or phrase that modifies the meaning of an adjective, verb, or other adverb, expressing manner, place, time, or degree. Ex: पर्याप्त (enough) |
| NOUN | Noun | A noun is a part of text that denotes a person, animal, place, thing, or idea. Ex: सेब (Apple) |
| NAME | Proper Noun | A name used for an individual person, place, or organization, spelled with an initial capital letter. Ex: ओमकार (Omkar) - [person name] |

*4) Translation:* All the segments whether words or partial simple sentences are translated individually. Words are translated based on its assigned POS tag referring to parallel Hindi-English dictionary [22]. If the Hindi word exist in the dictionary, the corresponding English word will be retrieved and tagged accordingly. On the other hand, if the word denoted as proper noun, it would be transformed to English by the Hindi-English transliteration. The words which represent many English words in translation, will be selected through learned SMT.

**Example 7.** *Input:-* 'विकास ने', 'विकास', 'किया'
*Translation:-* 'vikas', 'development', 'did'

*5) Rearrangement : Rule Based Translation:* In this phase we combine translated words and segments. Some rule based approaches are used to apply morphological modifications to the translated words i.e. add 'e', 'ed' or 'ing' after the verb, include "'s" with noun as apostrophe. Depending upon the assigned tags to the words, a matching grammar rule is selected for the given input. According to the selected sentence structure, required tense type of the sentence will be incorporated. And using that a proper linking verb i.e. am, is, are, was, were etc. are also required to be added to the sentence. Finally as per the matched grammar rule, words and other segments are rearranged and put in proper order. i.e some time "s" require after noun. We get final output after this phase.

V. RESULTS & DISCUSSION

*A. Data Description*

For word to word translation from Hindi to English, Hindi-English bilingual dictionary is utilized developed and maintained by CFILT, IIT-Bombay [22]. This bilingual dictionary contains 136,150 properly tagged Hindi words with their corresponding English words. To implement EBMT as well as to train statistical machine translation, "HindiEnCorp" release version 0.5 is used. This parallel corpus is made-up of 289,832 parallel sentences consists of 2.89 million of Hindi and 3.1 million English tokens as in table III [21].

Table III
HindiEnCorp corpora statistics

| Language Units | English | Hindi |
|---|---|---|
| Token | 2,898,810 | 3,092,555 |
| Types | 95,551 | 118,285 |
| Total Characters | 18,513,761 | 17,961,357 |
| Total Sentences | 289,832 | 289,832 |
| Sentences (word count ≤ 10) | 188,993 | 182,777 |
| Sentences (word count > 10) | 100,839 | 107,055 |

*B. Result*

Few translation results have been given in table IV. The input is given in the Devanagari script with UTF-8 encoding. First example is just a simple translation comprising of word-to-word conversion followed by grammatical rearrangement. Second example signifies the characteristic that how the proposed hybrid model capable of handling proper noun. Third example shows the robustness of translator to handle ambiguity in the sentence. Fourth one is the example of translating a sentence consists of phrases/idioms taken care of by EBMT part of the hybrid model. Last one is an example of a complex sentence translation.

Table IV
FEW EXAMPLES OF HINDI-ENGLISH TRANSLATIONS

| Hindi | English |
|---|---|
| भारत मेरा देश है | India is my country |
| मैं ओमकार विकास धारिया हूँ | I am Omkar Vikas Dhariya |
| विकास ने विकास किया | Vikas did development |
| ओंकार ने मुँह की बात छीनी | Omkar said what one was about to say |
| ऑटोरिक्शा दिल्ली में यातायात का एक प्रभावी माध्यम है | Autoriksha is an effective medium for journey in Delhi |

Word Error Rate (WER) based metric is used here to find the accuracy of the proposed approach. Fundamentally, WER is computed based on Levenshtein distance also known as the edit distance calculated through summing up the minimum no of insertions (I), deletions (D) and substitutions (S) applied to make a sequence similar to other. Consequently, accuracy of the MT system will be calculated through averaging the $Sent_{acc}$ on all the testing sentences.

$$WER = \frac{S + D + I}{N}$$

$$Sent_{acc} = 1 - WER$$

The proposed system's result is compared with Google, Microsoft BING and Babylonian translators on a set of 500 manually translated Hindi-English sentences which is made up of 150 complex sentences, 200 simple sentences, 75 idiom based sentences and 75 sentences with ambiguity. Google translator is basically an SMT type of translator which has to be learned on a big corpus for better efficiency and robustness. Likewise, MS-Bing is fundamentally based on both SMT and RBMT approaches for the translation. Similarly, Babylonian uses SMT and morphological operations to perform the translation. All in all, statistical learning plays a major role for a most accurate machine translator.

Table V
ACCURACY BASED ON DIFFERENT SENTENCE TYPE

| Sent Type / MT | Complex Sentence | Simple Sentence | Idioms | Sentences With Ambiguity |
|---|---|---|---|---|
| Proposed System | 73.17 | 90.12 | 96.55 | 94.74 |
| Google | 95.74 | 86.66 | 72.41 | 73.68 |
| MS-Bing | 85.36 | 83.33 | 34.48 | 68.42 |
| Babylon | 92.68 | 86.66 | 62.07 | 73.68 |

An statistical comparison of the proposed HMT based approach with Google, BING and Babylonian translators is presented in table V based on different sentence types. Analysis reveals that for simple sentences all MT systems show nearly same accuracy ranging between 83-90%. All system performance very well for simple and unambiguous sentence.

For Sentences includes idioms, proposed system gives more accurate output compare to other systems. The reason is the use of Example Based methods that is included in the proposed HMT system. The exact meaning of idioms is different than the actual meaning of word reside in it. Hence, the idioms could not be translated directly. We need the actual meaning or replacement of idioms in prior.

Sentences with ambiguity are handled by the SMT part of the proposed approach. Probability based statistical parameters, learned from the Hindi-English corpus, are used to resolve the ambiguity. To resolve the conflict between the proper noun and the dictionary word, various rules have been defined based on the morphological properties of sentence structure in Hindi.

## VI. CONCLUSION

In this paper, we demonstrate that hybrid model (HMT) of translation system is able to outperform many baseline translation systems based on EBMT, RBMT and SMT approaches individually. Currently the proposed system works only on four different types of tenses i.e. Simple Present Tense, Present Continuous Tense, Simple Past Tense and Past Continuous tense. In future the work can be extend on the remaining tenses. Currently it is not giving the good result on complex and multiple combine sentences it can be extended to be able to perform well for those sentences too after incorporating more complex grammatical rules under conformity with other modules of the system.